\documentclass[a4paper,man,natbib,floatsintext]{apa6}
\usepackage{lineno}
\usepackage{xcolor}
\usepackage{float}
\usepackage{amsmath}

\title{Characterizing the visual representation of objects from the child's view}
\shorttitle{The child's view of categories}

\author{Jane Yang$^1$, Tarun Sepuri$^1$, Alvin W.M. Tan$^2$, \linebreak Khai Loong Aw$^3$, Michael C. Frank$^2$, Bria Long$^1$}

\affiliation{
    $^1$Department of Psychology, University of California San Diego \\
    $^2$Department of Psychology, Stanford University \\
    $^3$Department of Computer Science, Stanford University
}

\begin{document}

\abstract{
  Children acquire object category representations from their everyday experiences in the first few years of life. 
What do the inputs to this learning process look like? 
We analyzed first-person videos of young children's visual experience at home from the BabyView dataset ($N$ = 31 participants, 868 hours, ages 5--36 months), using a supervised object detection model to extract common object categories from more than 3 million frames. 
We found that children's object category exposure was highly skewed: a few categories (e.g., cups, chairs) dominated  children's visual experiences while most categories appeared rarely, replicating previous findings from a more restricted set of contexts. 
Category exemplars were highly variable: children encountered objects from unusual angles, in highly cluttered scenes, and partially occluded views; many categories (especially animals) were most frequently viewed as depictions. 
Surprisingly, despite this variability, detected categories (e.g., giraffes, apples) showed stronger groupings within superordinate categories (e.g., animals, food) relative to groupings derived from canonical photographs of these categories. 
We found this same pattern when using high-dimensional embeddings from both self-supervised visual and multimodal models; this effect was also recapitulated in densely sampled data from individual children. 
Understanding the robustness and efficiency of visual category learning will require the development of models that can exploit strong superordinate structure and learn from non-canonical, sparse, and variable exemplars.  

}

\maketitle

\newcommand{\cat}[1]{\textsc{#1}}

\section{Introduction}

To form a visual category -- for example, an \cat{apple} -- young children need to extract the relevant visual features from diverse exemplars that may vary in shape, size, and even representational format (e.g., apple sauce, a line drawing of an apple, or a green apple on a table). Yet young children show evidence of achieving this computationally challenging feat quickly. Even four-month-olds can distinguish dogs from cats by silhouette alone \citep{quinn2001perceptual}. More extensive representations emerge gradually throughout the first and second years of life \citep{mareschal2001categorization, mandler1993concept, bergelson2017} as infants robustly recognize objects across different representational formats (e.g., line drawings), map them to labels, and even distinguish them from conceptually similar distractors \citep{bergelson2017, deloache1998grasping, zhu2025cross}.

How do young children come to understand their visual world so quickly? One key to answering this puzzle is to examine the actual input data for visual categorization, as any developmentally-plausible theory or model of visual learning must operate over these inputs. Videos taken from the infant perspective using head-mounted cameras  \citep{yoshida2008,aslin2009infants} suggest the child's view of the world is indeed dramatically different from that of adults \citep{yoshida2008}, and varies considerably as children learn to locomote on their own and interact actively with the objects, places, and people around them \citep{kretch2014,long2022a,smith2015,yoshida2008,aslin2009infants,franchak2011, sullivan2021}. Initial evidence analyzing the objects in the infant view suggests that children see some categories dramatically more than others, and experience highly variable, non-canonical views and exemplars of these categories \citep{clerkin2017real, long2021}.  

Children's naturalistic visual inputs may even be beneficial for learning. For example, children's ability to manipulate and rotate objects gives rise to diverse object views, supporting object perception in both young children and models \citep{soska2010systems, bambach2018toddler}. Additionally, both children and adults can learn visual category distinctions better when categories are sampled from skewed distributions versus uniform distributions \citep{lavi2021visual}. In one study, a small number of object categories were both pervasively present during mealtime for 8--10-month-olds (e.g., spoons, cups) and among infants’ first-learned words \citep{clerkin2017real, clerkin2022real}.

Despite this initial work, we still have an impoverished understanding of what the actual input data is for visual category learning. For example, how variable are the exemplars that children experience within and across categories? How often do young children tend to see certain categories (e.g., zoo animals) in real life versus as symbolic referents in storybooks or as toys \citep{deloache2004becoming}? And how consistent are children's visual experiences with object categories across individual households, even within relatively affluent and Western environments \citep{casey2022sticks}?

The answers to these questions have implications for both our theories as well as for models of visual category learning by constraining or expanding the set of relevant learning mechanisms. At present, modern neural network models for visual recognition -- including convolutional neural networks (CNNs) and vision transformers -- show immense promise as instantiations of statistical learning mechanisms that could operate over children's category learning inputs. Intriguingly, activations in these models to images of object categories successfully predict variation in both human object perception and neural responses to the same object categories \citep{yamins2014performance, conwell2024large}.  In particular, responses from object-selective cortex, deep neural networks, and human behavior exhibit a consistent geometry, with clustering within broad categories (e.g., animals versus inanimate objects) \citep{muttenthaler2021thingsvision, conwell2024large}. 

Yet despite their promise, these models are clearly receiving extraordinarily different kinds and amounts of visual data. Many models still require immense amounts of curated images to acquire useful representations relative to children \citep{ayzenberg2025fast,frank2023bridging,huber2023developmental} and learn best when given unrealistic amounts of explicitly labeled or captioned photographs \citep{radford2021learning, liu2023visual, schuhmann2022laion}. Thus, despite their correspondence with behavioral and neural data in adults, these models are likely quite far from being mechanistic models of children's visual category learning.

Here, we characterize the input data for category learning from the child's view by analyzing a large, longitudinal dataset of everyday experiences, with the goal of informing the learning mechanism that can explain how children learn visual categories so efficiently. 
To do so, we capitalize on new data and innovations in  computer vision to allow us to annotate these videos.
Until recently, existing open egocentric child video datasets have been relatively low-resolution videos with a narrow field of view from a handful of participants \citep{sullivan2021, long2021} and any analyses required laborious hand annotations. To overcome these challenges, we use the BabyView dataset \citep{long2024babyview}, an open dataset of egocentric video, whose first release contains 868 hours of data from $N$ = 31 children (release 2025.1, ages 5--36 months) from children living mostly in the United States. We also leverage new object detection and classification models \citep{wang2025yoloe, simeoni2025dinov3, radford2021learning} to identify common object categories across the entire dataset \citep{frank2023bridging,huber2023developmental}.

These new tools and data allow us to analyze the frequency, diversity, and similarity structure of the visual categories in the child's view, and to quantify how often children see real-life exemplars vs depictions across various formats -- as toys, drawings, or in media.  We compare exemplars of the categories in the child's view to images from THINGS \citep{stoinski2024thingsplus}, a curated dataset of images used widely in the vision science community. To do so, we use  visual representations for categories derived from the embedding spaces from both a self-supervised vision model \citep{simeoni2025dinov3} and a vision--language model \citep{radford2021learning}. We then use activations from these models to quantify the visual similarity of each category between BabyView and THINGS and the representational geometry that emerges in each dataset \citep{kriegeskorte2008representational}.

Using these data, we quantify how different the objects in the child's view are from the objects in curated datasets: We find that the distribution of objects in infants’ visual experience is indeed long-tailed, with some categories appearing dramatically more often than others, and that young children's experience contains substantial variation in viewpoint and across representational formats, with a high prevalence of depictions for many categories. Despite the variation, we observed that detections within superordinate categories were \textit{more} similar to each other in the child's view versus in a curated dataset. These results were recapitulated in individual data from densely sampled children, despite their idiosyncratic experiences. Overall, our findings both highlight the dramatic divergence between children's everyday experiences and curated datasets and open new avenues towards understanding how children's everyday category experiences scaffold their learning about the visual world.

\section{Results}

\begin{figure}
  \centering
  \includegraphics[width=0.99\textwidth]{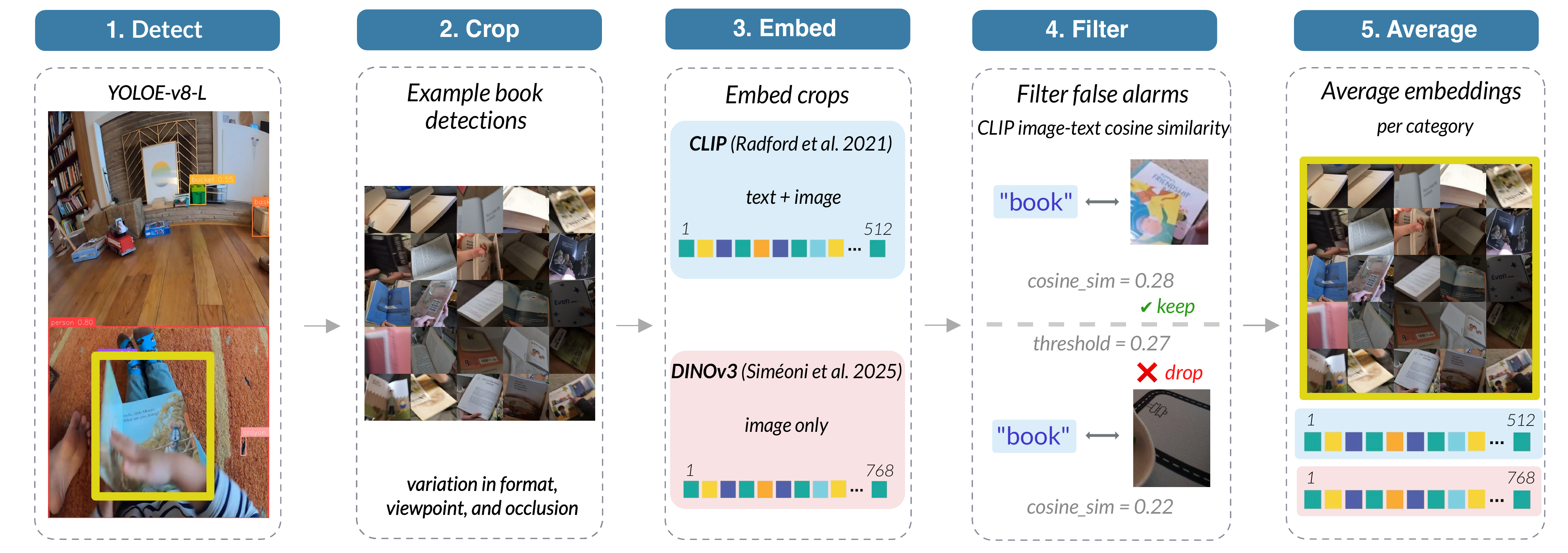}
  \caption{Overview of the automated detection pipeline. (1) YOLOE detects exemplars of CDI vocabulary categories in each frame; (2) bounding boxes are cropped, with example \cat{book} detections shown. (3) Each crop is embedded with both a vision--language model (CLIP) and a self-supervised vision model (DINOv3). (4) Detections are filtered by the cosine similarity between the CLIP image embedding of the crop and the CLIP text embedding of its predicted label, retaining only crops above a fixed threshold (0.27). (5) For category-level analyses, embeddings of retained crops are averaged within each category and dataset.}
  \label{fig:overiew}
\end{figure}

\subsection*{Identifying the object categories in the child's view}
We first extracted frames at 1 frame per second for a total of 3.68M frames from the 868 hours of the 2025.1 release of the BabyView dataset \citep{long2024babyview}. We then selected concrete nouns from the MacArthur-Bates Communicative Development Inventories (a commonly used questionnaire for measuring children's vocabulary; CDI) \citep{marchman2023} and performed category detection for $N$ = 129 categories using the YOLOE-v8-L model \citep{wang2025yoloe}; our final list of detected categories was based on iterative rounds of model validation (see Methods). We then filtered all object detections using CLIP, a vision--language model \citep{radford2021learning}, to decrease the probability of false alarms (see Supplemental Information (SI) 1.1). This pipeline resulted in a set of 2,994,667 detections for 129 categories across the entire dataset. 

For our full set of detections, we validated the precision of the pipeline detections by crowd-sourcing categorizations from human participants for a set of detections (100 per category) that were stratified across participants and videos. Using these data, we observed an average precision of .67 for our 129 categories ($SD$ = .22, see SI 1.2). For a strict set of detections, we constructed a set of human-validated, crowd-sourced detections for the 85 categories that had a precision greater than .60 to yield a ``gold'' set of 7,018 validated detections; in this subset, the average precision was .80 ($SD$ = .11, range = .61--.99). In both sets of detections, we observed that precision and detection frequency in the dataset were modestly correlated (full set of categories, $r$ = .34; strict set of categories, $r$ = .22; see Supplemental Figure~\ref{fig:precision_vs_detection_proportion_scatter}), suggesting that variation in observed frequencies is not a direct product of detection accuracy. 

As a final check on the robustness of our detection pipeline, we prompted a video question-answering (VideoQA) model, VideoLLaMA3 \citep{zhang2025videollama}, to detect \textit{all} objects in 10-second, contiguous chunks of the dataset \citep{sepuri2025characterizing}. We compared these detected categories and their frequencies with those extracted from our pipeline. The observed frequencies were relatively similar across these two distinct model pipelines ($r=.72$, $p<.01$, $N$ = 99 overlapping categories, see Supplemental Figure~\ref{fig:si_vqa_convergence}).
While object detection at scale in naturalistic egocentric video from the child's viewpoint is still a difficult challenge, this level of precision and our filtering pipeline (combined with robustness checks using a smaller set of human annotations) allows us to access a vastly larger amount of data about children's early visual experience than in prior work. 

\subsection*{The distribution of objects in infants' visual experience is long-tailed}

\begin{figure}
  \centering
  \includegraphics[width=0.99\textwidth]{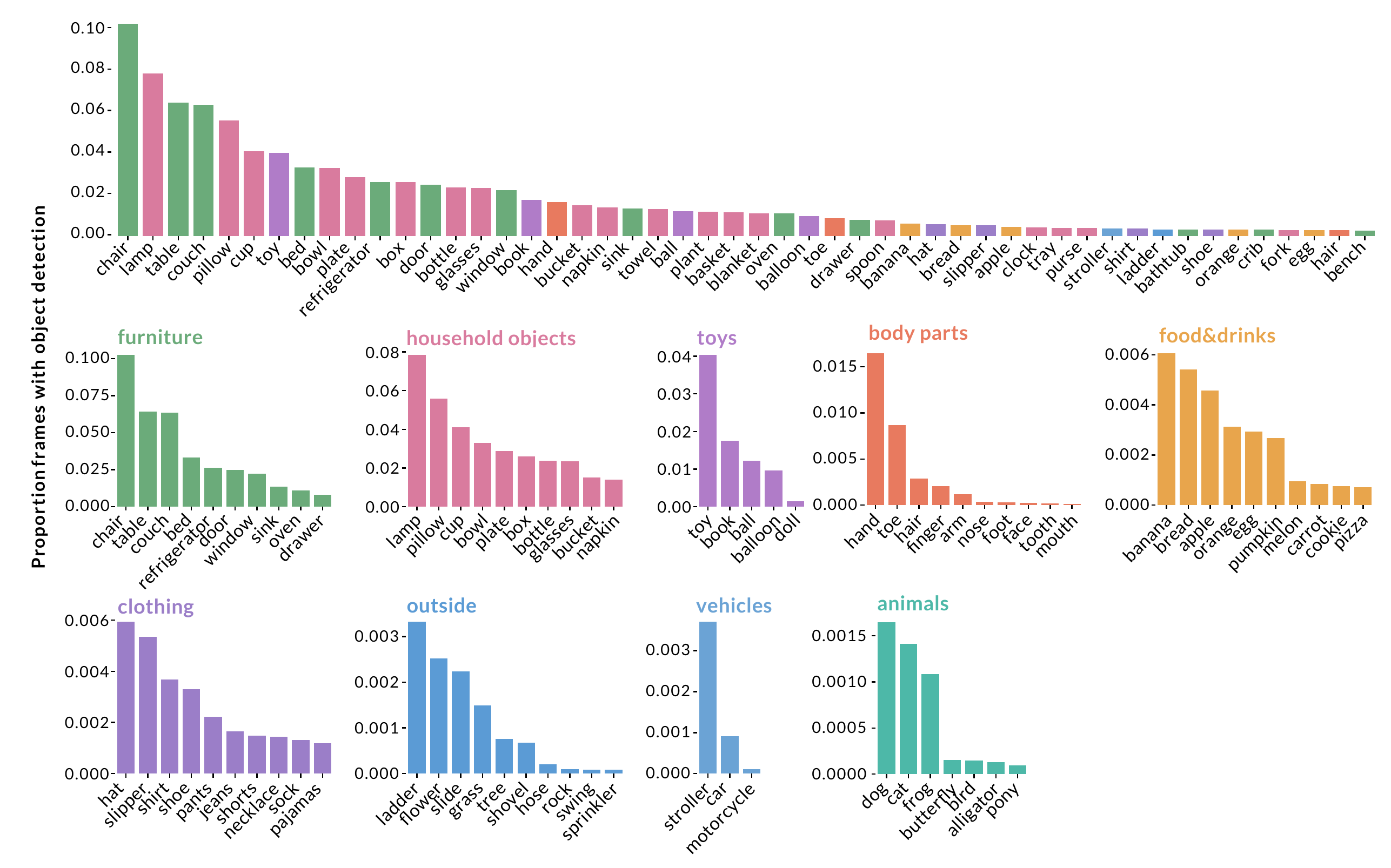}
  \caption{Long-tailed distribution of object categories in young children's everyday visual experience. Each bar shows the proportion of the 3.68M sampled frames containing at least one filtered detection of a given category. Top: the 50 most frequent categories overall, excluding \cat{person} and \cat{picture} (which dominate detections and further amplify the skew). Bottom: the most frequent categories within each CDI superordinate domain. Bars are colored by CDI domain.}
  \label{fig:longtail}
\end{figure}

We observed a skewed distribution of object categories in the child's view: a small number of object categories (e.g., \cat{chair, toy}) appeared very frequently while many others (e.g., \cat{penguin}) appeared less frequently, consistent with prior work \citep{clerkin2017real}. Figure~\ref{fig:longtail} shows the distribution of the top 50 most frequent categories that were detected in the dataset (excluding the most frequent \cat{person} and \cat{picture} detections, which only further amplify the skewed distribution); these detections are further broken down by their semantic category using categories from the CDI.

To quantify the shape of these distributions, we fit a power-law function to these frequencies, finding a power-law exponent of $\alpha$ = 1.93 across all 129 categories, comparable to the $\alpha$ = 2.44 reported by \citet{clerkin2017real} for hand-annotated object frequencies from egocentric videos of infants' mealtimes. Distributions were long-tailed even within each CDI category, with $\alpha$ = 1.23 (clothing), $\alpha$ = 1.98 (furniture), $\alpha$ = 1.93 (household objects), $\alpha$ = 1.68 (toys), $\alpha$ = 2.36 (body parts), $\alpha$ = 1.68 (food and drinks), $\alpha$ = 1.98 (outside), $\alpha$ = 3.13 (vehicles), and $\alpha$ = 1.69 (animals).

As a robustness check, we repeated all key analyses after restricting to human-annotated detections from 85 categories with human-validated precision and in the detections resulting from the VideoQA model pipeline (see Supplemental Figure \ref{fig:si_vqa_convergence}). The core pattern of results was unchanged in this high-precision subset or in the VideoQA model detections (see Supplemental Figure \ref{fig:si_valid85_longtail}). Overall, these results suggest that skewed distributions of object categories are a highly robust, general property of naturalistic visual environments as experienced by young children.

\subsection*{Object categories in the child's view vary in their similarity to canonical views in curated datasets}

Next, we examined how children experience individual exemplars and views of these categories. We compared the visual similarity between the average \cat{cup} that infants experienced to the average photograph of a \cat{cup} in a curated dataset -- THINGS \citep{stoinski2024thingsplus}, which is widely used to examine visual representations in both adults and children. To do so, we first took all filtered object crops in BabyView and all images for each category in the THINGS dataset (see Methods), and extracted embeddings from both a vision--language model \citep[CLIP;][]{radford2021learning} and a self-supervised vision model \citep[DINOv3;][]{simeoni2025dinov3} for all individual cropped images (see Methods) and averaged across exemplars.  In both cases, we computed the cosine similarity between the average category embeddings across the two datasets for 129 categories.\footnote{We anticipated that vision--language model (VLM) embeddings might better capture the higher-level similarity between the infant view and curated datasets, for two reasons. First,  VLMs are trained on inputs that vary in representational format (e.g., line drawings) and with semantic supervision (i.e., natural language captions for images) -- and, as noted, the child's view contains views of many depicted categories. However, we acknowledge that in all following analyses, that there is some circularity that could inflate any observed similarity between the child's view and curated dataset: Since all object detections were filtered for false alarms using embeddings from a VLM, our resulting set of filtered detections more already skewed towards the representational space of the VLM. Thus, the following results are likely to be an upper-bound on the true value, which we suspect might be somewhat lower if we were able to detect all possible \textit{valid} exemplars. Thus, we also utilized embeddings from a self-supervised vision model (DINOv3) to provide an independent estimate of the visual similarity between the detections in these datasets.}  

\begin{figure}[H]
  \centering
  \includegraphics[width=1\textwidth]{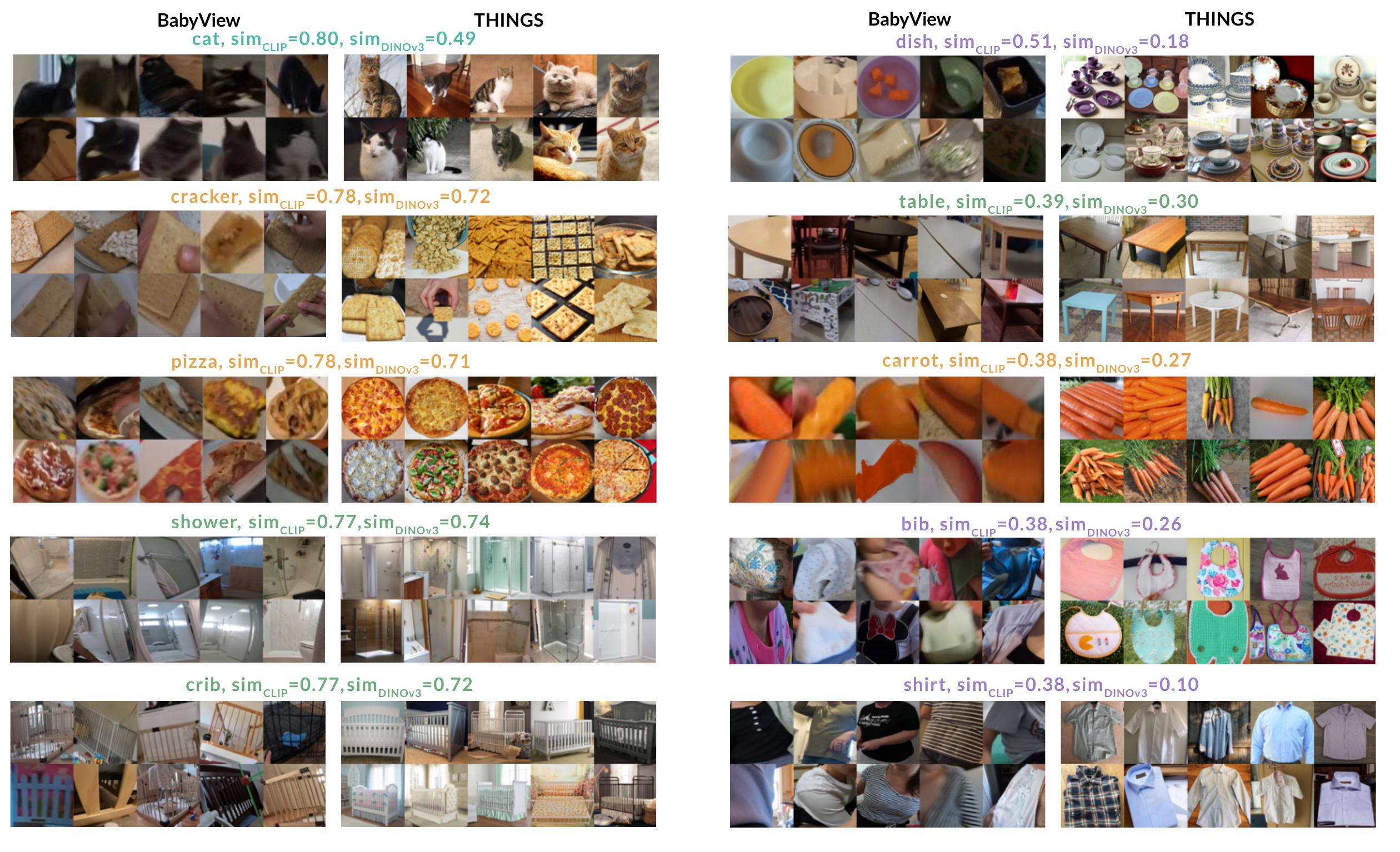}
  \caption{Example montages of detected exemplars from the child's view (BabyView) versus a curated dataset (THINGS), for categories with relatively high (left) and low (right) cross-dataset embedding similarity. Cosine similarity values between mean category embeddings are reported separately for CLIP and DINOv3; category labels are colored by their CDI superordinate domain.}
\label{fig:category_wise_cos_sim}
\end{figure}

We found that similarity values varied substantially across categories (average cosine similarity between BabyView vs THINGS, CLIP: range = .14--.80, mean $\cos(\theta)=0.61$; DINOv3: range = .04--.85, mean $\cos(\theta)=0.48$; see Figure~\ref{fig:category_wise_cos_sim}). There was relative agreement across the two feature spaces on the question of which categories were more or less similar (Pearson's correlation between DINOv3 vs. CLIP category-wise similarity values: $r = 0.70$, $p < .01$). There were some differences, however, particularly for categories with diverse exemplars, where the DINOv3 embeddings appeared to capture more of this variability. For example, \cat{dish} in BabyView tended to be close up views of plates vs. collections of dishes in THINGS (see Figure \ref{fig:category_wise_cos_sim})---here, the DINOv3 embeddings were more sensitive than CLIP to this viewpoint and exemplar variability. 

However, there are many ways in which the objects in the child's view differ from curated datasets; this variation does not fall neatly along domain boundaries. In our analysis, no broad category domain (animate objects, furniture, toys) was consistently more or less similar to its curated counterpart. The differences we observed not only appears to reflect differences in visual format but also in the range of viewpoints and contexts through which children encountered particular categories; objects in children's everyday experiences are often seen from non-canonical angles, under variable lighting conditions, and often partially occluded. 

\subsubsection{Variation across visual formats}
Qualitatively, we observed that the objects in the child's view spanned a wide range of visual formats: birds were experienced as photographs, alligators were experienced as illustrations and line drawings, and many children had toy versions of cars, trucks, and trains. However, curated datasets rarely capture this variation in visual format: for example, all of the exemplars in the THINGS dataset are photographs of realistic exemplars. Accordingly, we observed that categories with high variation in visual format across their exemplars (e.g., real objects, toys, and depictions) tended to show lower correspondence between BabyView and THINGS in DINOv3.

\begin{figure}[H]
  \centering
  \includegraphics[width=1\textwidth]{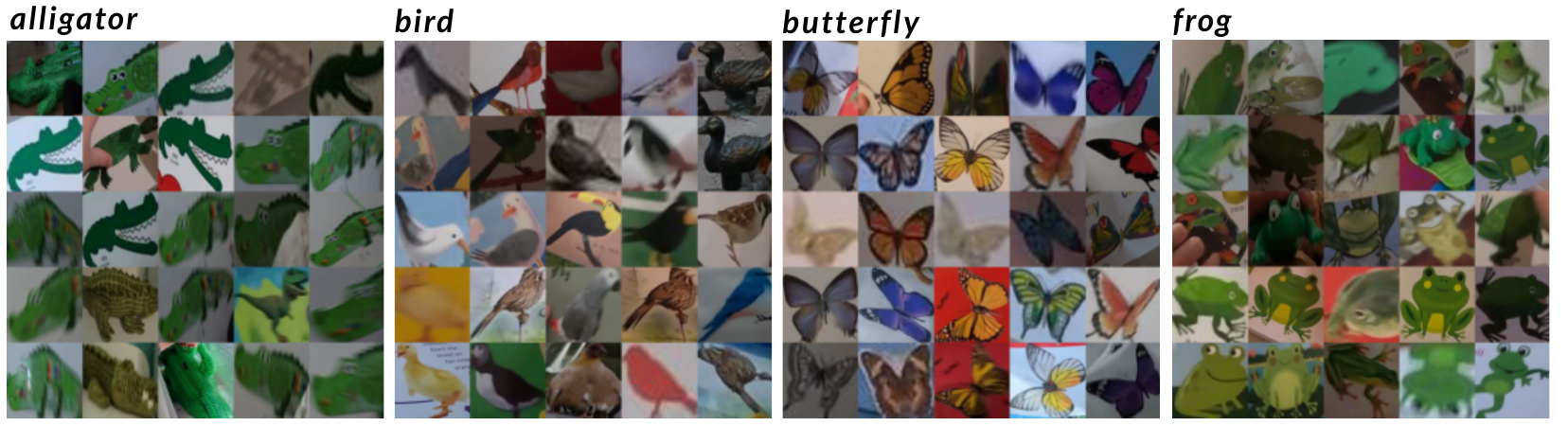}
  \caption{Human-validated exemplars from selected animal categories, illustrating the prevalence of depictions (toys, drawings, photographs, and screen media) in young children's everyday experience. }
  \label{fig:animal_montages}
\end{figure}

We additionally conducted a manual annotation of the detected exemplars for animals to analyze how frequently children experienced real-life exemplars, given the observed range of visual formats they were observed in. Trained annotators simply classified exemplars as belonging to a real-life vs. a depicted viewpoint (including photographs, media, stuffed animals, and line drawings). These analyses revealed that depictions constituted a substantial proportion of the animal exemplars, with some variation across categories. For example, 100\% of the exemplars in \cat{pony} were depictions (98\% in \cat{butterfly}, 94\% in \cat{bird}, see examples in Figure~\ref{fig:animal_montages}).  Conversely, animals that children might experience as household pets -- dogs and cats -- had overall lower proportions of depictions (23\% in \cat{dog}, and 5\% in \cat{cat}) These results suggest that for many categories, the majority of children's learning experiences in the home -- and perhaps even outside of it, e.g., in daycare -- will consist of depicted exemplars.

\subsection*{Between-category representational geometry in young children's object experiences}
\begin{figure}
  \centering
  \includegraphics[width=.99\textwidth]{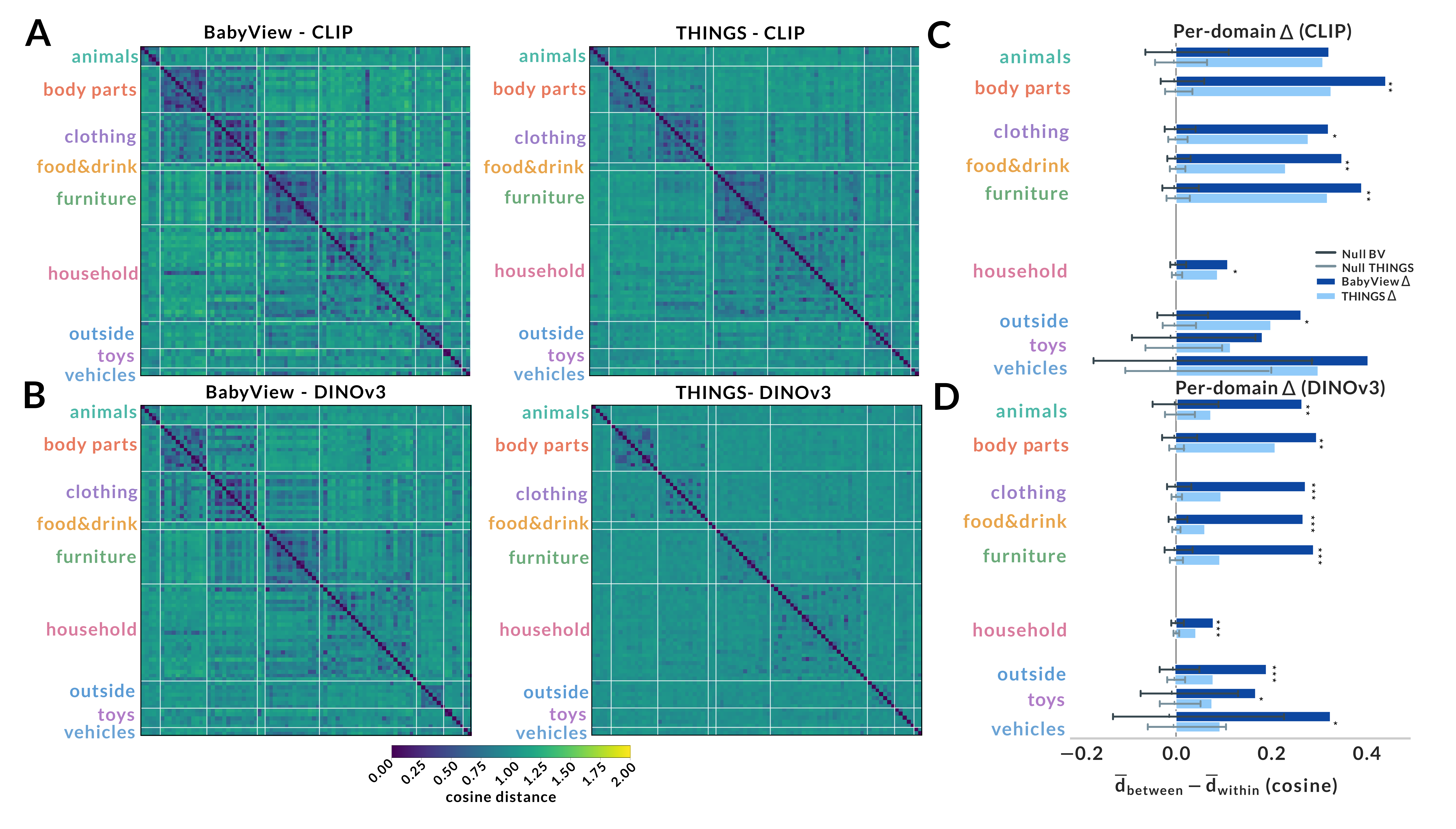}
  \caption{Between-category representational geometry and CDI-domain clustering in BabyView versus THINGS (129 categories). (A) BabyView--CLIP versus THINGS--CLIP RDM comparison. (B) BabyView--DINOv3 versus THINGS--DINOv3 RDM comparison. In both RDM panels, matrices are 129$\times$129 pairwise cosine distances between category centroids, ordered by CDI superordinate domain; darker diagonal blocks indicate stronger within-domain similarity. 
  (C) Per-domain cluster strength in CLIP. (D) Per-domain cluster strength in DINOv3. For panels C--D, bars show $\Delta_d=\overline{d}_{\mathrm{between}}-\overline{d}_{\mathrm{within}}$ for BabyView and THINGS, with permuted null intervals shown for each dataset/domain. Across most domains, clustering is stronger in BabyView than in THINGS.}
  \label{fig:bv_things_rdm}
\end{figure}

We used Representational Similarity Analysis (RSA) to compare between-category geometry in BabyView versus THINGS (see Methods). For each dataset, we computed a 129$\times$129 representational dissimilarity matrix (RDM) from pairwise cosine distances between category-mean embeddings, then correlated the strict lower triangles between datasets. Cross-dataset agreement was moderate in both feature spaces (CLIP: $\rho = 0.55$, $p < .01$; DINOv3: $\rho = 0.40$, $p < .01$; Figure~\ref{fig:bv_things_rdm}). 
Although moderate in absolute size -- as expected given differences in viewpoints, context, and noise -- these effects indicate reliable agreement in \emph{which} category pairs are relatively similar vs.\ dissimilar across datasets.

When we examined the RDMs from both BabyView and THINGS, we observed coherent groupings for the CDI superordinate semantic categories: animals, body parts, clothing, large household objects, and small household objects clustered together. For example, the average embedding of \cat{cat} was more similar to other animals than to clothing or household objects. We found this to be the case when we constructed RDMs using embeddings from either CLIP or DINOv3, with some variability (see Figure \ref{fig:bv_things_rdm}).

Remarkably, these broad category clusters appeared overall \textit{stronger} in data from the child's view vs. in THINGS. To quantify this CDI-domain cluster strength, we constructed  a between-minus-within distance statistic $\Delta_d$ for each superordinate domain ($\Delta_{d}^{\mathrm{BV}}$ and $\Delta_{d}^{\mathrm{TH}}$; see Methods), where larger values indicate stronger within-domain clustering. As a random baseline, we then shuffled CDI labels across categories while preserving domain counts (using the same shuffled labeling in BabyView and THINGS on each permutation draw (see Methods; SI 1.7). We first examined activations from CLIP: here, we found that clustering in BabyView was strongest for body parts ($\Delta_{d}^\mathrm{BV} = 0.440$), vehicles ($0.403$), and furniture ($0.390$), and weakest for household objects ($0.108$), a broad and heterogeneous domain. THINGS showed the same rank ordering but weaker separation overall (body parts $\Delta_{d}^\mathrm{TH} = 0.325$, furniture $0.318$, vehicles $0.298$; household objects $0.087$). After within-model false discovery rate correction, the BabyView $>$ THINGS domain contrast was significant in 6 of 9 domains for CLIP and in all 9 of 9 domains for DINOv3 (See Figure~\ref{fig:bv_things_rdm}C \& D). Collectively, these findings indicate that CDI superordinate structure is present in both datasets and is generally more pronounced in the child's naturalistic visual input.


\subsection*{Idiosyncratic experiences across children support similar category structures}

\begin{figure}
  \centering
  \includegraphics[width=.99\textwidth]{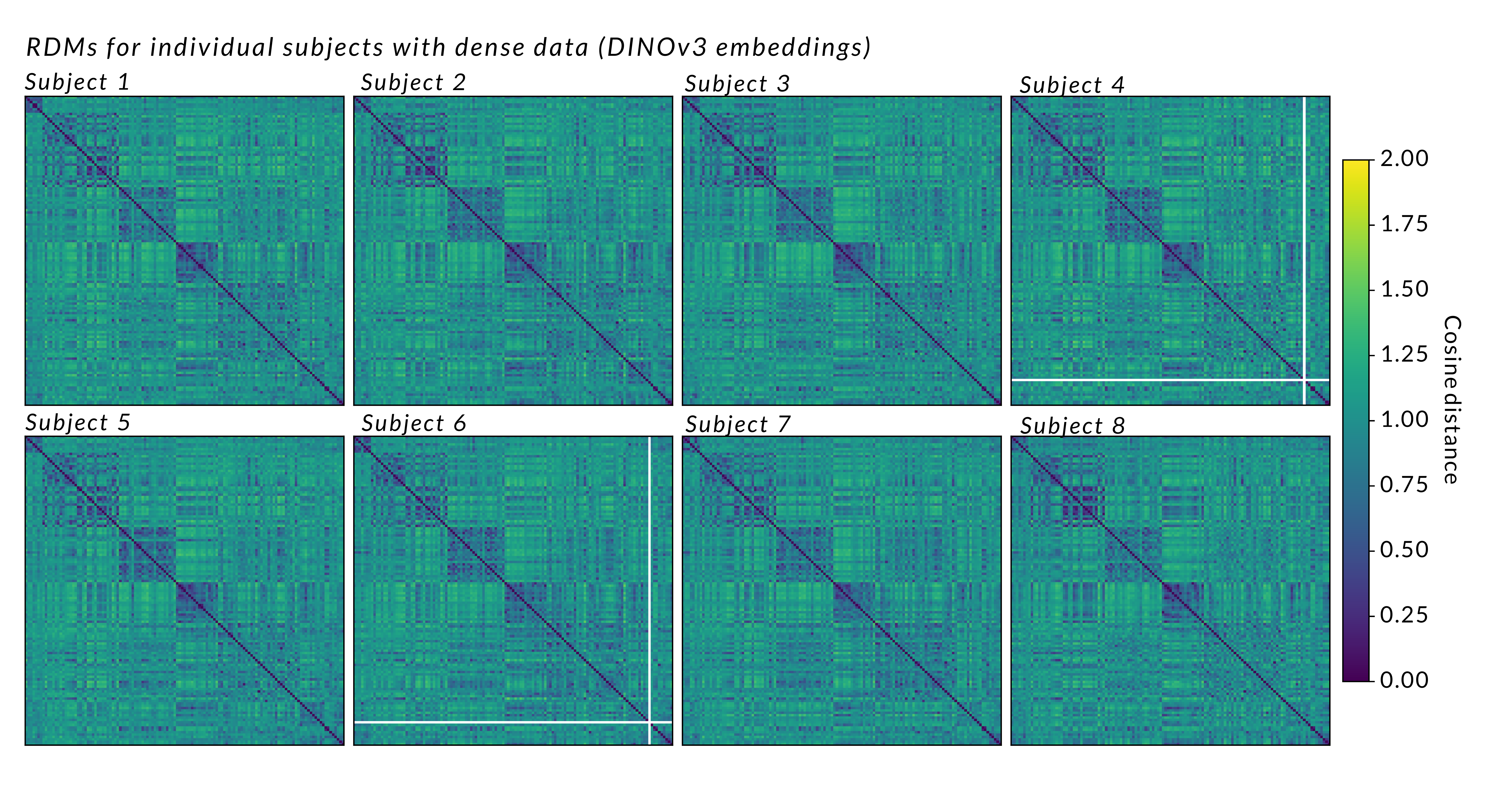}
  \caption{Individual-family RDMs (DINOv3 embeddings, all 129 categories) for the eight participants with the densest recordings. Categories are ordered by CDI superordinate domain, matching Figure ~\ref{fig:bv_things_rdm}. Thick white lines indicate missing categories. Darker values indicate visually similar category pairs (low cosine distance); lighter values indicate dissimilar pairs. The same broad superordinate structure visible in the aggregate RDM is recoverable in each individual family, despite idiosyncratic home environments.}
  \label{fig:DINOv3_top8_rdm}
\end{figure}

Every child has their own unique home environment. Does the category structure we observed in an average across families reflect consistent properties of each individual child's visual experience? To examine this question, we constructed separate RDMs for the eight children with the densest recording data ($>$ 32.5 hours of data per family, range = 32.5--91.5 hours, ages 7--29 months). Despite the unique visual environments of each family, the broad organizational structure observed in the aggregate was remarkably consistent. Figure \ref{fig:DINOv3_top8_rdm} shows the individual RDMs in DINOv3 embedding space, highlighting again that coherent clusters for animals, body parts, and large household objects were identifiable in each family's RDM (categories are in the same order as Figure \ref{fig:bv_things_rdm}).
Statistically, we found that these individual RDMs were relatively similar to each other in embeddings from both models (average between-subject RDM correlation for CLIP: $r=0.776$, $SD=0.039$; average between-subject RDM correlation for DINOv3: $r=0.787$, $SD=0.032$; see Figure \ref{fig:DINOv3_top8_rdm}). Thus, this category structure appears across the families in our sample in both embeddings from a vision--language model and a purely visual self-supervised model. Overall, these findings indicate that the category structure documented in our aggregate analyses is not merely an artifact of averaging across diverse inputs but is a property that arises from the visual experiences of individual children.


\section{Discussion}

How do young children experience visual categories?  We quantified the statistical properties of young children's everyday visual experiences with object categories in a large dataset of egocentric videos using innovations in computational models and human annotations. Within our sample, children saw some objects dramatically more than others: a heavily skewed frequency distribution dominated by common household objects, with substantial variability at the exemplar level, consistent with the distributions observed in prior work \citep{clerkin2017real, clerkin2022real, long2021}. 

Further, we found stronger clusters of individual categories within superordinate groupings in the child's view vs curated datasets -- highlighting that children experience \textit{more} redundancy within superordinate groupings than is present in third-person curated photographs.  Furthermore, we found that this between-category geometry was highly stable across the everyday experiences of the individual children in our dataset. 

First, these results constrain theories of visual category learning by extending findings from prior work. Our results reinforce the idea that early category learning operates over repeated encounters with a small set of objects -- children's cups, toys, household objects, a handful of real-life animals, and many depictions \citep{clerkin2017real, clerkin2022real, long2021}. These skewed distributions may be both ubiquitous in natural environments \citep{newman2005power} and in fact beneficial for visual statistical learning in adults \citep{lavi2021visual}.

However, our results also highlight how objects may be experienced in dramatically differ ways across broad category domains. For small, handheld objects, repeated, self-generated views are likely beneficial for building robust 3D object representations. 
Children who have more experience sitting and rotating objects with their hands tend to succeed on a 3D object completion task \citep{soska2014postural}, and variance in these self-generated views tend to predict vocabulary growth \citep{slone2019self}. 
However, many object categories in the infant view do not have this property and cannot be rotated or manipulated by children.  Indeed, children must navigate around or on top of large, immovable objects (chairs, slides, tables) \citep{konkle2013tripartite}. And other categories (e.g., many animals) may be mostly experienced as depictions: as ``child-directed'' visuals that emphasize their diagnostic features. Quantitative theories of children's category learning will need to expand to incorporate these very different kinds and quantities of experience across broad category domains. 

Indeed, a key finding of this work is that children's experiences of objects cluster \textit{within} superordinate categories more strongly than in curated datasets: that is, children experience more variable exemplars than is captured in curated datasets, but this additional variability is consistent \textit{within} superordinate clusters. The redundancy within broad, superordinate categories may provide a wedge for children to parse the ``blooming buzzing confusion'' into meaningful categories that can be mapped to words in their native language \citep{bergelson2017}. Indeed, this intersects with a foundational idea in cognitive development that children acquire global representations (e.g., animacy) before basic-level representations \citep{mandler1988cradle, mandler2000global, quinn2000emergence}. This is consistent with an emerging computational perspective that training computational models with these broad category distinctions -- as opposed to basic-level or subordinate level labels -- is sufficient for the emergence of human-like object representations \citep{mehta2026extremely}.

Our results also have implications for research that seeks to emulate child-like learning in machine learning models. While the ``data gap'' between humans and computational models of visual learning has so far been quantified in terms of hours of data or number of images \citep{orhan2021much, frank2023, huber2023developmental, ayzenberg2025fast} our work suggests that part of the reason for this gap may be a fundamental difference not only in the quantity but in the \textit{types} of objects that children experience. 
Indeed, while an emerging literature has trained unsupervised models on images and videos from head-mounted camera data \citep{zhuang2021, orhan2021much, orhan2024learning, long2024babyview}, they have only found mixed success compared with models trained on canonical training datasets such as ImageNet \citep{deng2009imagenet}. Our findings thus point towards  a fundamental algorithmic gap between humans and models: the fact that models require such radically different learning inputs suggests they are not simply running a noisier or slower version of the same algorithm; modeling research may need to seek new algorithms fundamentally different in kind. For example, model architectures that can leverage the relational structure between categories might better succeed at learning from these non-canonical, sparse exemplars. Our work does not prescribe a particular algorithmic solution, but contributes a ``cognitive target'' and suggests properties that future benchmarks should capture.

There are several limitations to these findings. First, the YOLOE detections were far from perfect, highlighting a need for annotation tools that are suited to the child's perspective; while we found convergent results on the frequencies of different categories using detections from a different model class \citep[a VQA model; see][]{sepuri2025characterizing}, we are likely missing certain categories or exemplars in our analysis.  Our detection and filtering approach is conservative -- that is, by passing all detections through a vision--language model, we likely removed from exemplars or viewpoints of object categories that happen to be even more unusual -- and these viewpoints may be relatively challenging for YOLOE to start with. However, we still observed the same pattern of results when we included all possible detections and categories vs. filtered detections. Thus, our results may represent an upper-bound on the similarity between children's everyday experience and curated datasets, and we suspect that a complete characterization of the child's view would reveal even more divergence. 

Second, our sample size is relatively small for some of our densest longitudinal analyses, limiting generalizability. Other aspects of the dataset also limit generalizability: for example, the dataset captures primarily indoor environments due to privacy constraints, potentially missing important aspects of visual experience that occur outdoors and in daycare or caregiving settings. Finally, this dataset captures the experiences of infants from a specific geographic and cultural context, and the generalizability of these statistical patterns to other populations remains an open question \citep{henrich2010weirdest}.

Overall, this work provides empirical data about the visual input available to developing minds by quantifying the object frequencies, exemplar variability, and category relationships in children's everyday, naturalistic settings. The combination of skewed frequencies, variable exemplars, and stable relational structure characterizes a learning environment that differs markedly from typical laboratory or computational settings. Children's ability to build categorical knowledge from this input highlights the need for developmental theories that can account for how everyday learning occurs across messy, embodied, naturalistic contexts.

\section{Acknowledgments}
We gratefully acknowledge the families who participated in the BabyView Dataset. We are grateful to Mira Mateo, Dora Deng, and Jason Yang for assistance with video annotation and ground-truth coding.  This work was funded by an NIH R00HD108386 grant to B.L., by a grant from Schmidt Futures, by a gift from Meta, by the Stanford Center for the Study of Language and Information John Crosby Olney Fund, and by the Stanford Human-Centered AI Initiative (HAI) Hoffman-Yee grant program.

\section{Methods \& Materials}

\subsection*{Dataset}
Data for this study came from the 2025.1 release of the BabyView dataset \citep{long2023, long2024babyview}, consisting of 868 hours of egocentric video recorded using head-mounted cameras from 31 infants (aged 5--36 months) during their everyday activities.  Data are available at \url{https://www.databrary.org/volume/1882}.
We sampled frames at 1 frame per second, resulting in 3.68 million frames for analysis.


\subsection*{Automatic Object Detections using YOLOE}
We detected objects in each frame using the YOLOE-v8-L model \citep{wang2025yoloe}. Frames from egocentric videos are out of domain for YOLOE. To validate YOLOE's performance on our dataset, we employed two complementary manual annotation strategies: ground-truth annotations to assess detection completeness and recall, and model corrections to assess precision and label accuracy. First, the detected categories were restricted to align with concrete nouns in the MacArthur-Bates Communicative Development Inventories (CDI) vocabulary items \citep{marchman2023} ($N$ = 295 words). We included all detections above YOLOE's default confidence threshold of .25. We excluded the frequent \cat{person} and \cat{picture} detections from subsequent analyses as they had heterogeneous referents. Detections of \cat{person} commonly included body parts, such as toes and ears. Detections of \cat{picture} commonly included detections of photo frames and drawings. Additionally, we excluded any categories that had less than 100 total exemplars, yielding a set of 129 categories.

\subsubsection{Ground-truth annotations} 
Next, two trained coders were each independently given 58 randomly sampled frames (116 frames in total). Annotators were instructed to annotate all possible categories within the CDI vocabulary list ($N$ = 295), as well as add new labels if a label were the most suitable to an object but not within the existing CDI vocabulary list. We obtained a limited number of ground-truth frames only because it was labor-intensive. For each object, annotators were also asked to add an overlapping bounding box of \cat{picture} or \cat{toy} to the object if it were a depiction of an object (e.g., ``a picture of a dog''). Overall, this initial round of annotations revealed that YOLOE detections were reasonably accurate for $N=163$ categories; however, many categories were very infrequently present, with sometimes only a few exemplars present per category in the frames themselves.

\subsubsection{Detection filtering}
We aimed to exclude false alarms for each category that could increase noise in our similarity analyses. To create a cleaner subset of high-confidence true positive detections for the THINGS comparison, we implemented a CLIP-based filtering procedure. For each YOLOE detection, we extracted the cropped image region within the bounding box and passed it through CLIP's image encoder (ViT-B/32; \citealp{radford2021learning}), yielding a 512-dimensional image embedding. We also encoded the predicted category label (e.g., \cat{cup}, \cat{dog}, \cat{chair}) using CLIP's language encoder, yielding a corresponding text embedding. We then computed the cosine similarity between the image embedding and its associated label embedding. This similarity score reflects how well the visual content of the cropped region matches the semantic meaning of the predicted category label according to CLIP's learned visual--linguistic alignment.

Detections were retained only if their image-text cosine similarity exceeded a threshold of .27, which was determined through qualitative examination of detections at various thresholds. At this threshold, we observed that retained detections were predominantly true positives with clear, recognizable instances of the labeled category, while filtered detections often contained either incorrect objects, extremely partial views, heavy occlusions, or empty/ambiguous regions where YOLOE had hallucinated an object. For example, a detection labeled \cat{dog} showing a clear, frontal view of a dog would typically achieve similarity scores above .30, while a false alarm showing a piece of furniture mislabeled as \cat{dog} would score well below .26.  This filtering procedure removed a substantial majority of the original YOLOE detections, with filtering rates varying by category. Categories with clearer visual appearance and less ambiguity (e.g., \cat{ball}, \cat{book}) had lower filtering rates, while categories with more variable appearance or greater susceptibility to YOLOE errors (e.g., \cat{toy}, \cat{food}) had higher filtering rates.  

At this threshold, CLIP filtering removed 83.06\% of raw YOLOE detections overall (retaining 2,994,667 of 17,674,191 detections). Filtering rates varied substantially across categories. Because this filter is conservative, it likely removes some true positives with atypical appearance, partial visibility, or unusual viewpoints, potentially biasing retained detections toward more prototypical exemplars.

\subsubsection{Model detection verification} We assessed the precision of model detections by asking human annotators to verify filtered detections via a crowd-sourced annotation task. We obtained three independent human ratings verifying our model detections for each of 100 exemplars from 129 different categories (12,900 crops in total). To construct this validation set, we sampled from CLIP-filtered detections with additional diversity constraints: categories were required to have at least 100 eligible exemplars across a large spread of subjects and videos (see SI 1.2). 

This model verification strategy thus allows us to compute the accuracy of the filtered detections for each category. However, this annotation task was considerably more challenging than annotating in the objects in the full-view frames taken from the video, due to the loss of the surrounding contextual information. In our main analyses, we thus report the main results for all $N$ = 129 categories, as well as for a stringent filtered subset of $N$ = 85 categories that have a precision > .60. This dual reporting strategy is intended to preserve comparability while quantifying sensitivity to false-alarm-prone categories.

\subsection{Object representations}
We investigated the category structure of the BabyView dataset using embedding representations and representational similarity analysis \citep{kriegeskorte2008representational}.

\subsubsection{Object embeddings}
We embedded all object crops from the filtered dataset using the CLIP image encoder (ViT-B/32; \citealp{radford2021learning} as well as the DINOv3 image encoder (Vit-B/16; \citealp{simeoni2025dinov3}). 
CLIP embeddings represent more semantically aligned representations (due to the joint language--image pretraining), whereas DINOv3 embeddings represent more vision-focused representations (due to image-only self-supervised pretraining).

\subsubsection{Category-level summaries}
For each included category we summarized detection frequencies from the filtered frame-level detections (1\, frame per second sampling) and grouped categories by CDI superordinate domain labels used throughout the manuscript. For BabyView versus THINGS alignment at the \emph{exemplar} level, we compared mean category embeddings using cosine similarity (CLIP and model-matched DINOv3), as reported in the Results.

\subsubsection{CLIP-threshold sensitivity analysis}
To test whether representational results depended on the CLIP image--text operating threshold, we reran the $N=129$ aggregate pipeline at thresholds $t\in\{.26,.27,.28\}$ while holding category inclusion and category order fixed; see Supplemental Figure~\ref{fig:si_clip_threshold_sensitivity_valid129}. For each threshold, BabyView category centroids were recomputed from the retained detections; THINGS centroids were unchanged. We then recomputed the RDM-based diagnostics used in the main analyses (BabyView--THINGS Spearman correlation in CLIP and DINOv3 spaces, and within-dataset CLIP--DINO RDM agreement), and also tracked total retained detections within the included category scope. This analysis quantifies the tradeoff between stricter filtering (fewer retained detections) and stability of the recovered representational geometry.

\subsection{Representational dissimilarity matrices (RDMs) and cross-dataset alignment}
\subsubsection{Construction and Spearman correlation}
We asked whether the \emph{pattern} of between-category distances in BabyView matches THINGS when both datasets use the same $N=129$ categories in the same order. For each embedding model (CLIP and DINOv3), we averaged exemplar embeddings within category (separately for BabyView and THINGS), computed all pairwise cosine distances between category centroids, and formed one symmetric RDM per dataset. We then vectorized each RDM's strict lower triangle (one entry per unique category pair) and computed Spearman's rank correlation $\rho$ between BabyView and THINGS vectors. Two-sided $p$-values came from the standard Spearman test on those paired vectors.

\subsubsection{Label-shuffle null for cross-dataset $\rho$}
To test whether cross-dataset $\rho$ depends on true category alignment, we implemented a \emph{single-sided label-shuffle null} (separately for CLIP and DINOv3). BabyView was fixed. On each of $n_{\mathrm{perm}}=5{,}000$ draws, we sampled one random permutation of the 129 category labels and applied it to THINGS \emph{rows only} (columns unchanged). We then vectorized the strict lower triangle of this row-permuted THINGS matrix and recomputed Spearman's $\rho$ against the fixed BabyView lower-triangle vector. Empirical two-sided $p$-values were computed as $(k{+}1)/(n_{\mathrm{perm}}{+}1)$, where $k$ is the number of null draws with $|\rho_{\mathrm{null}}|\geq |\rho_{\mathrm{obs}}|$.

\subsection{CDI superordinate alignment in embedding space}
We quantified CDI superordinate structure using a \emph{per-domain} statistic in each fixed RDM, $\Delta_d=\overline{d}_{\mathrm{between}(d)}-\overline{d}_{\mathrm{within}(d)}$, computed separately for BabyView and THINGS. Inference/benchmarking for this analysis used label shuffling across categories with domain counts preserved (5,000 draws), with the same shuffled domain assignment applied to both datasets on each draw.

\subsubsection{Domain-level cluster strength}
Let $D(i)$ denote the CDI domain of category $i$ and $d_{ij}$ the pairwise cosine distance between categories $i$ and $j$ in a fixed RDM ($i<j$ throughout). For each domain $d$, we compute
\[
\Delta_d \;=\; \overline{d}_{\,\mathrm{between}(d)} - \overline{d}_{\,\mathrm{within}(d)},
\]
where $\overline{d}_{\,\mathrm{within}(d)}$ is the mean pairwise distance among categories assigned to domain $d$, and $\overline{d}_{\,\mathrm{between}(d)}$ is the mean distance from categories in $d$ to categories outside $d$. We denote this quantity as $\Delta_{d}^{\mathrm{BV}}$ for BabyView and $\Delta_{d}^{\mathrm{TH}}$ for THINGS. Larger $\Delta_d$ indicates stronger superordinate clustering for that domain (between-domain distances exceed within-domain distances by a larger margin). Because domains reuse the same underlying pairwise distances, these per-domain summaries are not independent; we therefore interpret them primarily as a profile of \emph{where} clustering is stronger vs.\ weaker across the CDI taxonomy.

\subsubsection{Marginal label-shuffle intervals for $\Delta_{d}$}
To provide a visual null benchmark for each $\Delta_d$, we estimated \emph{marginal} label-shuffle intervals. On each draw, we held both RDMs fixed and shuffled CDI domain labels across categories (category identities and distances unchanged), while preserving exact domain counts. The same shuffled domain assignment was applied to BabyView and THINGS in parallel on that draw. We recomputed $\Delta_{d}^{\mathrm{BV}}$ and $\Delta_{d}^{\mathrm{TH}}$ for all domains, repeated this over 5,000 draws, and took the 2.5th and 97.5th percentiles as the central 95\% interval. Under this null, a label such as ``animals'' refers to a random subset of categories of matching size, not the true animal set. These intervals are used as a descriptive visual benchmark, not as a multiplicity-corrected decision rule across domains.
For clarity of presentation, we visualize this same 2.5--97.5\% marginal null interval in equivalent styles (e.g., capped whiskers/error bars in the supplement, shaded bands where used elsewhere). All styles are constructed from identical per-domain permutation draws and differ only in rendering.

\subsubsection{Agreement across domains between datasets}
We quantified agreement between BabyView and THINGS in the \emph{pattern} of domain strengths by Spearman rank correlation between the paired vectors $\{\Delta_{d}^{\mathrm{BV}}\}_{d}$ and $\{\Delta_{d}^{\mathrm{TH}}\}_{d}$ within each embedding model (two-sided $p$-values from SciPy's \texttt{spearmanr}).

\subsubsection{Per-domain BabyView-versus-THINGS inference and FDR control}
For each CDI domain $d$ and embedding model, we quantified the directional contrast $\Delta_{d}^{\mathrm{BV}}-\Delta_{d}^{\mathrm{TH}}$ and computed a one-sided permutation $p$-value for the hypothesis BabyView $>$ THINGS using the shared count-preserving label-shuffle draws described above. Because this yields a family of nine per-domain tests per model, we controlled false discoveries within each model using Benjamini--Hochberg false discovery rate correction across domains and report corresponding $q$-values in supplemental outputs.

\subsection{Analyses of individual-family RDMs (dense-recording subsample)}
To examine whether aggregate geometry reflects stable structure within individual children's homes, we identified the eight families with the densest usable recording time under the same filtering and category inclusion rules used for the aggregate RDMs. Operationally, we ranked families by total hours of analyzed video contributing to the included category set and selected the top eight families that each exceeded 32.5 hours (range 32.5--91.5 hours; child ages at recording 7--29 months). For each family we computed centroid RDMs in CLIP and DINOv3 embedding spaces using the same pipeline as for the aggregate BabyView RDM. We then quantified pairwise similarity of these family-specific RDMs by Spearman correlation over the strict lower triangle (restricted to category pairs present for both families in a given comparison) and summarized the distribution of between-family correlations (mean and SD across pairs), providing empirical bounds on how consistent superordinate structure is across idiosyncratic visual environments.

\bibliography{references}


\appendix

\section{Supplemental Information}

\subsection*{SI 1.1. CLIP filtering pipeline and false-alarm control}

To reduce category-specific false alarms from open-vocabulary YOLOE detections, we applied a CLIP-based image-text consistency filter to every detection crop. For each predicted object, we computed cosine similarity between the CLIP image embedding of the crop and the CLIP text embedding of the predicted category label, and retained detections only when similarity exceeded .27.

At this threshold, CLIP filtering removed 83.06\% of raw YOLOE detections overall, retaining 2,994,667 of 17,674,191 detections. Filtering rates varied by category, with ambiguous categories showing higher rejection rates than visually specific categories. This conservative filter improves precision for downstream representational analyses, while likely excluding some true positives with atypical viewpoints, heavy occlusion, or low-resolution appearance.

\subsection*{SI 1.2. Crowdsourced annotation protocol and derivation of the human-validated subset}

We constructed a crowdsourced validation set by sampling 100 detection crops per category from the CLIP-filtered YOLOE detections (129 categories; 12,900 regular crops in total in the annotation manifest). Annotators ($N=12$) were shown 25 images at a time and asked to ``Please identify all invalid detections of [cat]''; each trial included a pre-specified invalid image as an attention check. Annotators who failed more than 20 attention checks were replaced. Sampling was constrained to increase diversity and reduce over-representation from repeated clips: categories were required to have at least 100  exemplars after CLIP filtering, at least 8 unique subjects, and sufficient spread across videos in the dataset to include at most 2 exemplars from a single video when forming the annotation set (average video length = 510.77s).

Each sampled crop received 3 independent human ratings (per-file table: $N_\textup{raters}=3$). For each crop, we computed an error rate from the crowd responses and converted this to crop-level precision ($\textup{precision}=1-\textup{error rate}$); class-level precision was then computed by averaging crop-level precision within category. Across the 129 categories, mean class precision was .669 (reported as 0.67 in the main text; $SD$ = 0.22).

For high-precision robustness analyses, we defined a stringent subset of 85 categories using a fixed threshold of precision $>$ .60 at both levels: (i) category-level precision and (ii) exemplar-level precision, intersected with the pre-specified category list (N = $129$) and the sampled manifest rows. This procedure yielded 85 categories and 7,018 unique human-validated exemplar crops (the $\sim$7k set used in supplemental robustness analyses).


\subsection*{SI 1.3. Convergence with video question-answering model detections}

To validate our automated pipeline, we prompted a video question-answering (VideoQA) model, VideoLLaMA3 \citep{zhang2025videollama}, to detect all objects in 10-second, contiguous chunks of the dataset. We utilized the same chunks of the dataset as presented in \cite{sepuri2025characterizing}, sampling every third chunk for a total of $N$ = 289.33 hours. We prompted the model with the phrase ``This a video from the point-of-view of a camera mounted on a child's head. Strictly return a list detailing each object, animal and person present in this video, comma separated like so: \lq ball,tennis racquet,person,sofa\rq.'' We compared detections both for the full set of $N$ = 129 categories and the stringent high-precision subset of $N$ = 85 categories. For the full set of categories, we found 99 categories that overlapped between the VideoQA model and our automated detections. The 30 missing categories were all small objects (for example, \cat{toe} and \cat{orange}). One reason these detections were missing in longer videos could be that they were not salient across longer 10-second chunks of the dataset and occurred for shorter durations.

We found strong correlations in the observed frequencies of overlapping categories across both the full set of categories ($r$ = .72, $p$ < .01, $n$ = 99) and the stringent high-precision subset of categories ($r$ = .69, $p$ < .01, $n$ = 70). Additionally, we observed a skewed distribution of objects. We fit a power-law function to the frequencies in the high precision set of categories, finding a power-law exponent of $\alpha$ = 2.57 across the overlapping 70 categories, comparable to the $\alpha$ = 2.44 reported by \cite{clerkin2017real} and the $\alpha$ = 1.93 found with our YOLOE model pipeline.

\begin{figure}[H]
  \centering
   \includegraphics[width=0.6\textwidth]{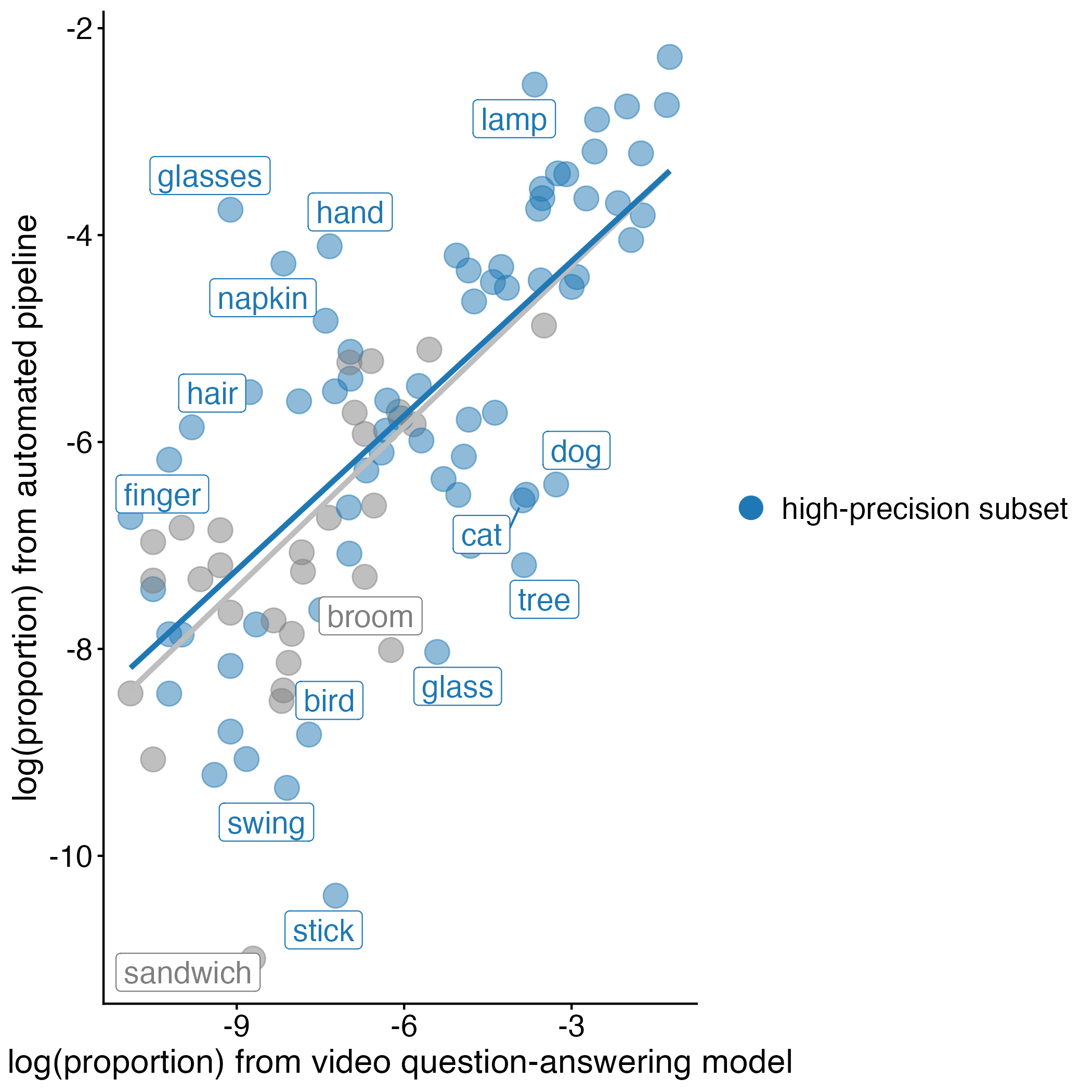}
   \caption{Convergence between the automated YOLOE+CLIP pipeline and an independent video question-answering pipeline (VideoLLaMA3). Each point is a category, plotted by its log proportion of detections in each pipeline. Gray points and regression line: full set of overlapping categories ($N$ = $99$); blue points and line: high-precision human-validated subset ($N$ = $70$). Thirty categories detected by YOLOE but not returned by the VQA model are excluded.}
   \label{fig:si_vqa_convergence}
\end{figure}


\subsection*{SI 1.4. Long-tailed category distribution in the high-precision subset ($N$ = 85)}

We observed the same long-tailed structure when restricting analyses to the high-precision human-validated subset ($N$ = 85). Fitting a power-law function to category frequencies yielded an exponent of $\alpha=1.92$ across the 85 included categories. Semantic-level fits (when sufficient categories were available) also showed long-tailed structure, with estimated exponents of $\alpha=1.97$ (animals), $\alpha=2.35$ (body parts), $\alpha=1.23$ (clothing), $\alpha=2.00$ (furniture), $\alpha=1.61$ (household objects), $\alpha=2.30$ (outside), and $\alpha=1.68$ (toys). Food \& drinks and vehicles each contained only two categories in this subset and were therefore too sparse for stable semantic-level power-law fitting.

\begin{figure}[H]
  \centering
  \includegraphics[width=0.9\textwidth]{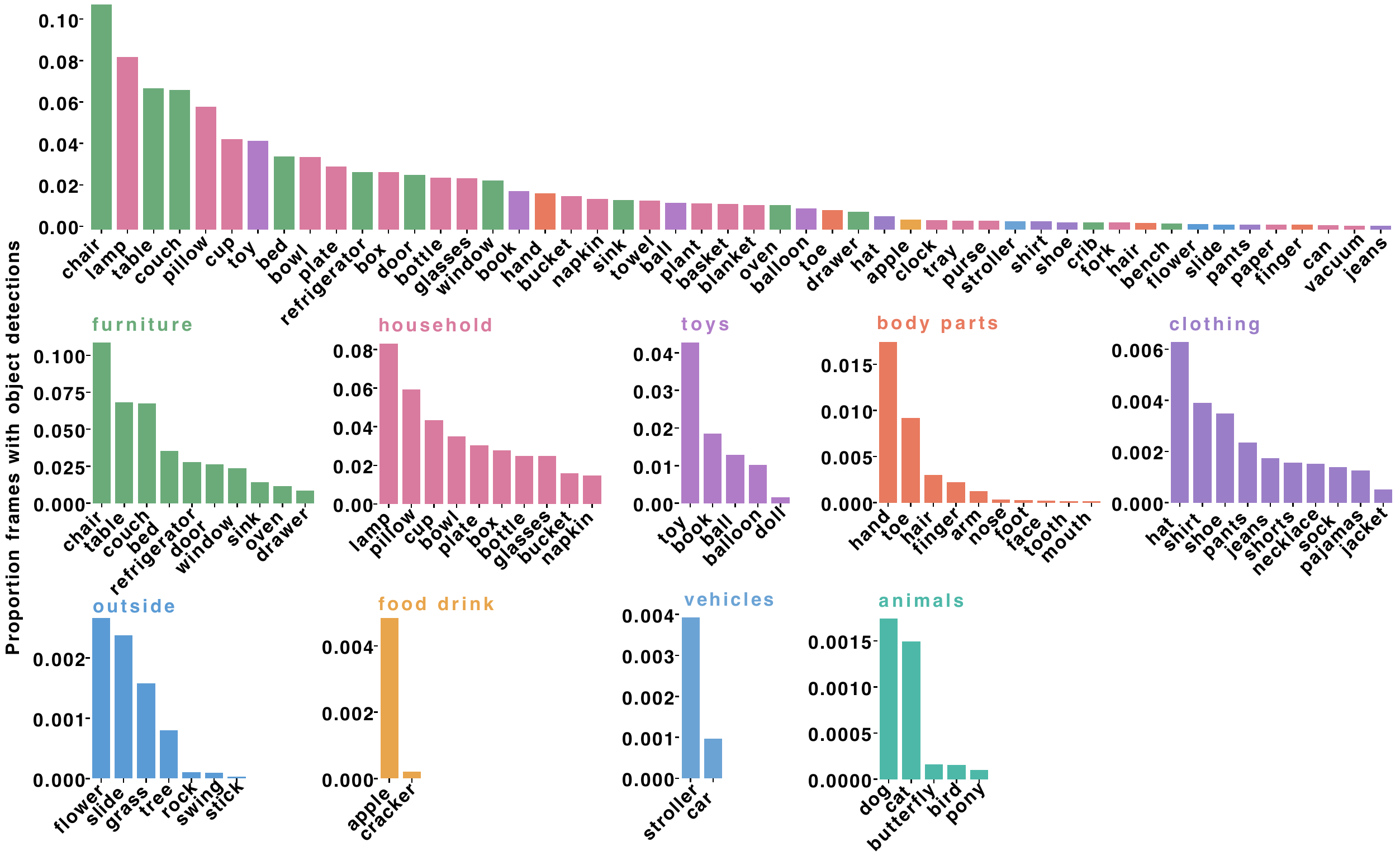}
  \caption{Long-tailed distribution of object categories in the high-precision human-validated subset ($N$ = $85$). As in Figure ~\ref{fig:longtail}, each bar shows the proportion of the 3.68M sampled frames containing at least one filtered detection. Top: all 85 categories, ranked by frequency. Bottom: category frequencies within each CDI superordinate domain. The skewed shape observed in the full set of 129 categories is preserved in this stricter subset (overall power-law exponent $\alpha = 1.92$).}
  \label{fig:si_valid85_longtail}
\end{figure}

\begin{figure}[H]
  \centering
  \includegraphics[width=0.9\textwidth]{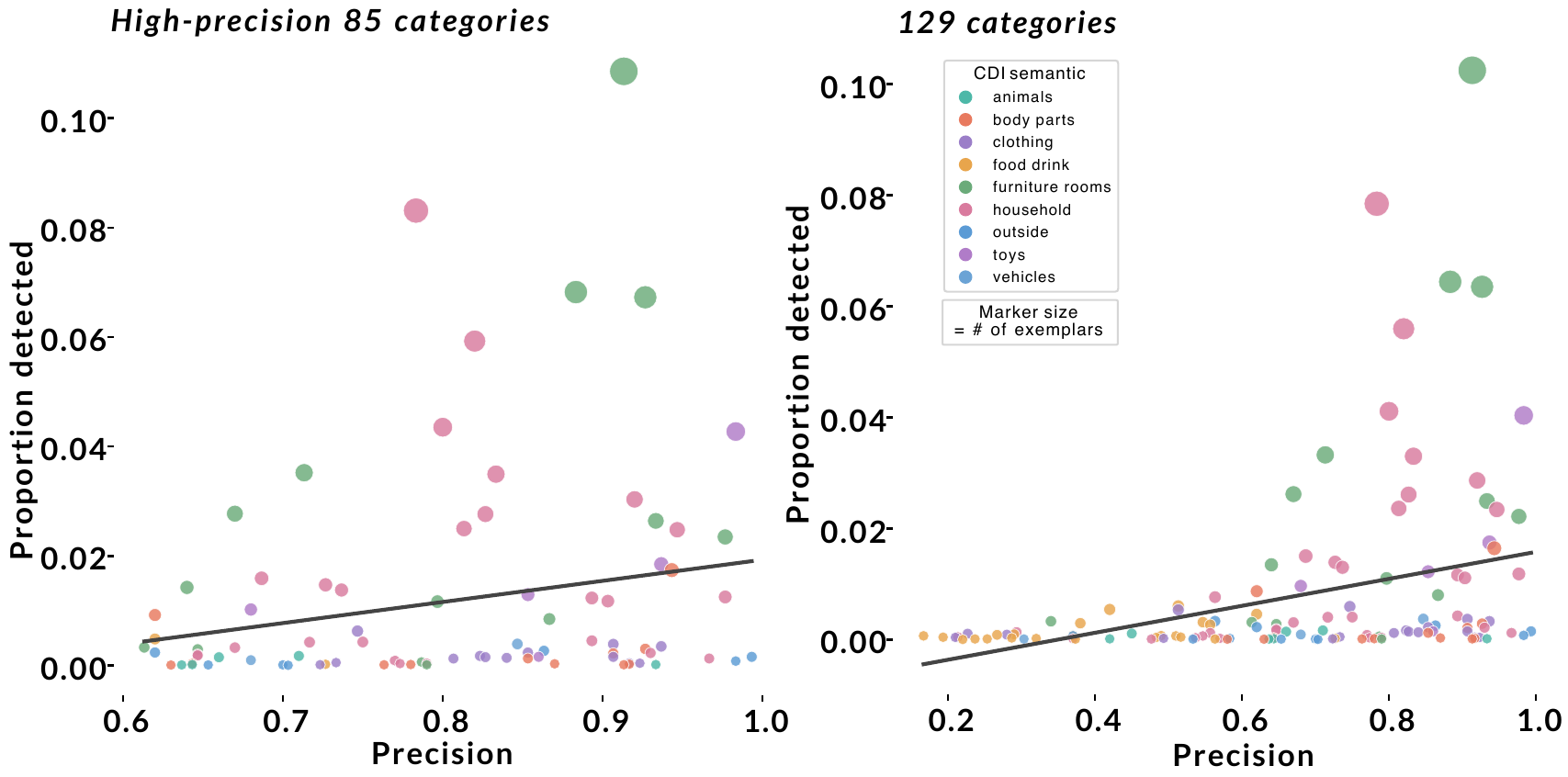}
  \caption{Relationship between detection frequency and human-validated precision. Each point is a category, plotted by its crowdsourced precision (x) against its proportion of CLIP-filtered detections within the analysis set (y). Left: high-precision subset ($N$ = 85); right: full set ($N$ = 129). Points are colored by CDI superordinate domain and sized by total detection count; lines show ordinary least-squares fits. Frequency and precision are only modestly correlated, indicating that the most frequent categories are not disproportionately driving precision estimates.}
  \label{fig:precision_vs_detection_proportion_scatter}
\end{figure}

\subsection*{SI 1.5. Category-wise BabyView--THINGS similarity distribution ($N$ = 85)}

For category-wise similarity (BabyView vs. THINGS), values showed substantial heterogeneity across the 85 categories in both embedding spaces. In CLIP space, cosine similarity ranged from .32 to .82 (mean = .62, median = .63, $SD$ = .11). In DINOv3 space, cosine similarity ranged from .07 to .84 (mean = .50, median = .49, $SD$ = .18).

\begin{figure}[H]
  \centering
  \includegraphics[width=0.9\textwidth]{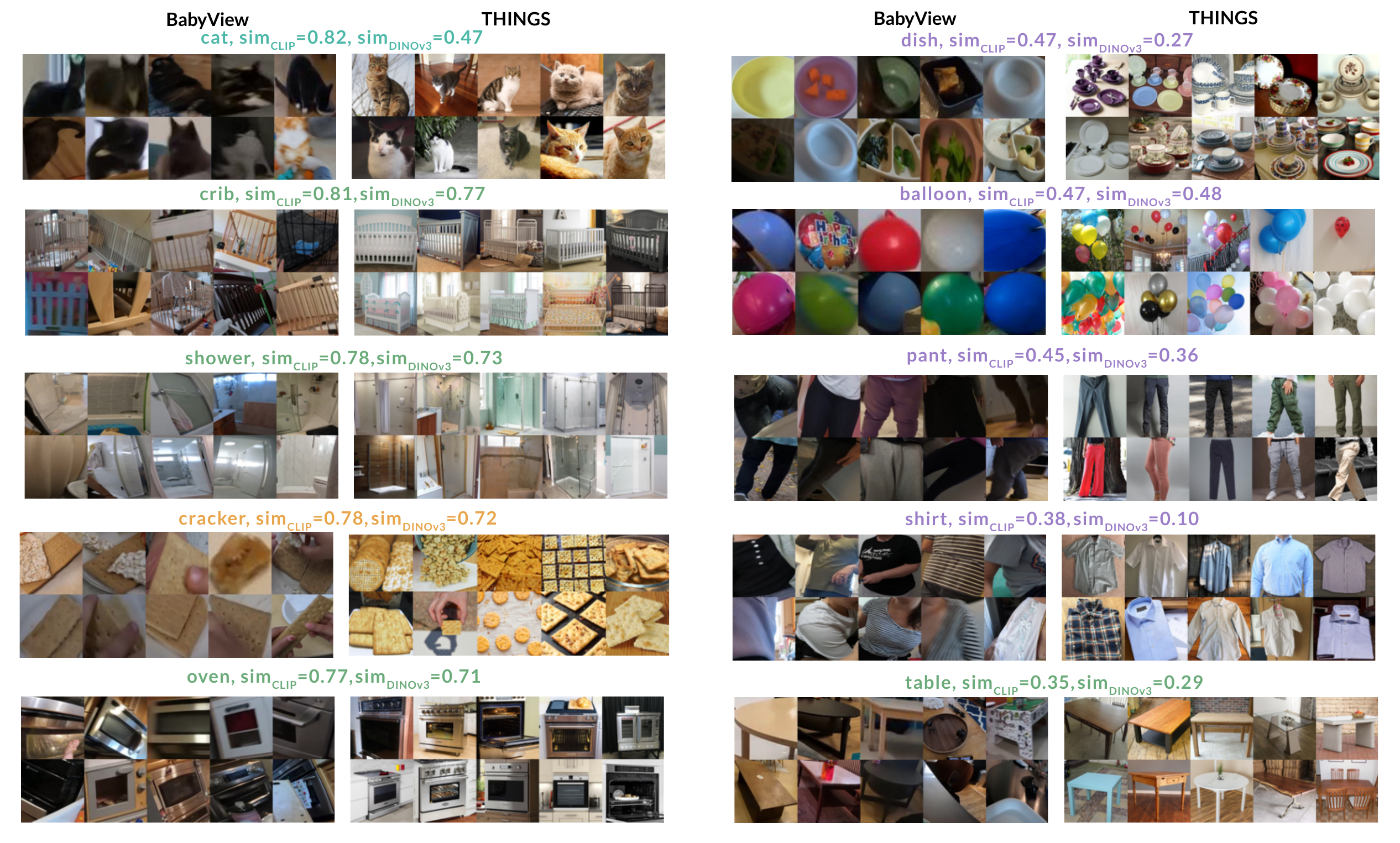}
  \caption{Example exemplar montages for the high-precision subset ($N$ = 85), shown as in Figure ~\ref{fig:category_wise_cos_sim}. Left: categories with relatively high cross-dataset similarity between BabyView and THINGS; right: categories with relatively low similarity. Cosine similarity values are reported separately for CLIP and DINOv3.}
  \label{fig:si_valid85_category_similarity}
\end{figure}

\subsection*{SI 1.6. Representational geometry convergence with THINGS ($N$ = 85)}

Using the high-precision human-validated subset ($N$ = 85), we observed moderate convergence between representational structure in BabyView and THINGS. In CLIP space, the BabyView-vs-THINGS RDM correlation was Spearman's $\rho=.57$ ($p<.01$); in DINOv3 space, the corresponding correlation was Spearman's $\rho=.46$ ($p<.01$).

\begin{figure}[H]
  \centering
  \includegraphics[width=0.95\textwidth]{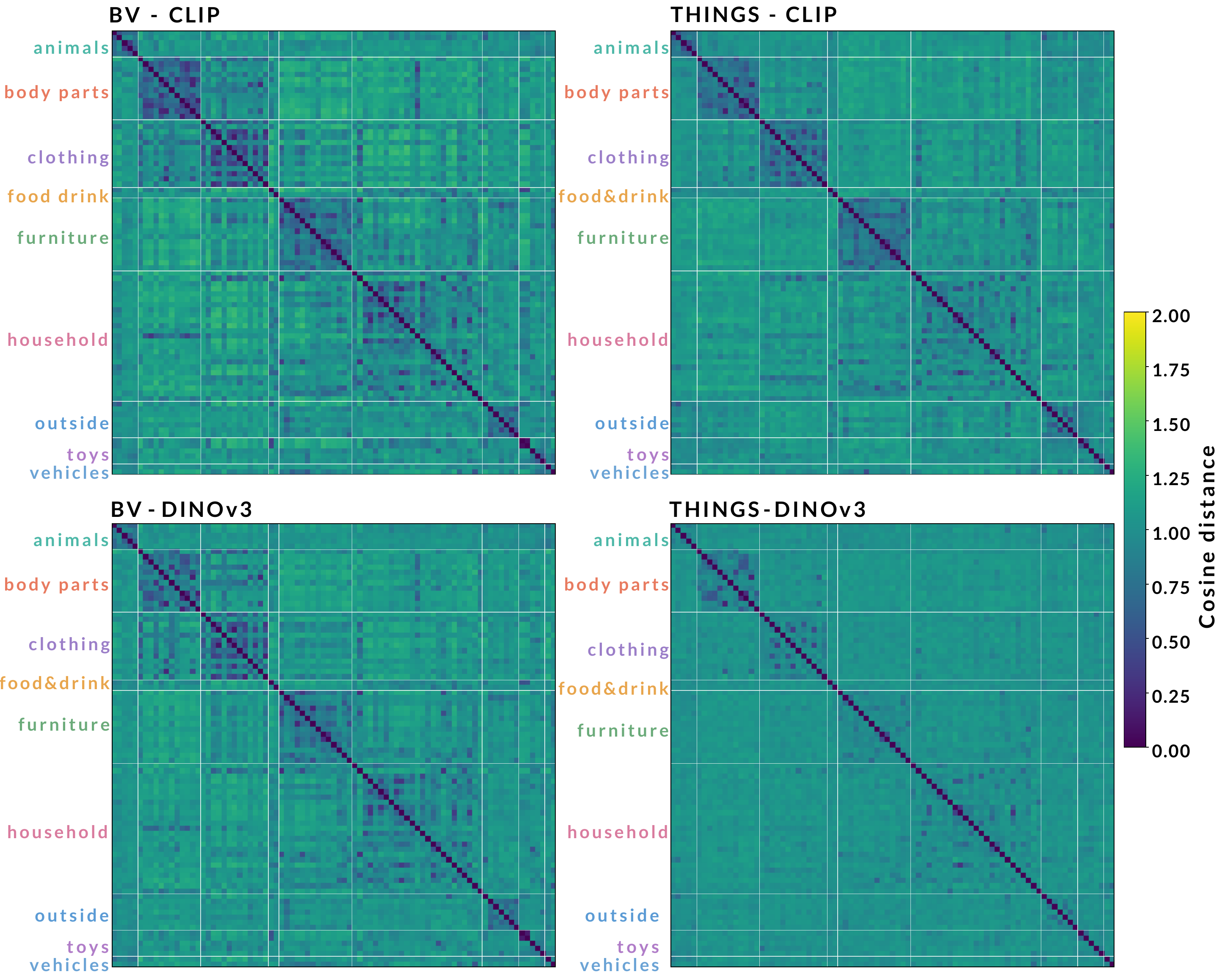}
  \caption{Between-category representational geometry for the high-precision subset ($N$ = 85 categories). Each panel shows an 85×85 RDM of pairwise cosine distances between mean category embeddings, with categories ordered by CDI superordinate domain. Top row: CLIP; bottom row: DINOv3; left column: BabyView; right column: THINGS. The aggregate pattern from Figure 4 is preserved: superordinate clusters are visible in both datasets and are more pronounced in BabyView (CLIP: Spearman's $\rho = .57$; DINOv3: Spearman's $\rho = .46$; both $p < .01$).}
  \label{fig:si_valid85_rdm}
\end{figure}

\subsection*{SI 1.7. Within- versus between-cluster separation for CDI semantic groups}

Procedures for domain-level $\Delta_{d}$ and its label-permutation null ($n_{\mathrm{perm}}=5{,}000$) are specified in \emph{Methods} (main text). Here we restate the implemented shuffling procedure for clarity and show the null interval as whiskers for readability.

For each CDI domain $d$ and model, we compute $\Delta_d=\overline{d}_{\mathrm{between}}-\overline{d}_{\mathrm{within}}$ separately for BabyView and THINGS. On each permutation draw, we keep both RDMs fixed and shuffle CDI superordinate labels across categories (equivalently: random category-to-domain reassignment), while preserving the exact count of categories per domain. The same shuffled label vector is then applied in parallel to BabyView and THINGS for that draw. We recompute $\Delta_d$ under this shuffled labeling and repeat over 5,000 draws. Whiskers denote the 2.5th and 97.5th percentiles of these per-domain null draws (central 95\% interval); bars denote observed $\Delta_d$. Thus, whiskers are a visual benchmark for ``how large $\Delta_d$ is under label-randomized domains'' and are not a multiplicity-corrected decision rule across domains.

Domain-level $\Delta_{d}$ was largest for body parts, vehicles, and furniture/rooms and smallest for household in both embeddings; Spearman rank correlation of $\{\Delta_{d}^{\mathrm{BV}}\}$ vs.\ $\{\Delta_{d}^{\mathrm{TH}}\}$ across domains was CLIP $\rho=0.88$, $p=0.0016$, and DINOv3 $\rho=0.73$, $p=0.025$ ($k=9$ domains), indicating similar domain ordering with larger separation in BabyView for most domains.

For per-domain directional contrasts (BabyView $>$ THINGS), we computed one-sided permutation $p$-values and applied Benjamini--Hochberg FDR correction across the 9 domains within each model. In CLIP, 6/9 domains survived FDR ($q<.05$: body parts, clothing, food/drink, furniture/rooms, household, outside), while animals, toys, and vehicles did not. In DINOv3, all 9/9 domains survived FDR ($q<.05$), indicating a more pervasive BabyView $>$ THINGS separation profile in that embedding space.


\subsection*{SI 1.8. Agreement between CLIP and DINOv3 pairwise geometry}

For each dataset (BabyView and THINGS) and each category set, we formed the symmetric RDM of pairwise cosine distances between mean category embeddings, then correlated the off-diagonal entries of the CLIP RDM with those of the DINOv3 RDM (same category order; CLIP detection filter threshold $.27$ as elsewhere). Table~\ref{tab:si_clip_dino_rdm} summarizes Pearson and Spearman correlations ($n(n-1)/2$ pairs). Agreement was much stronger for BabyView than for THINGS, indicating that CLIP and DINOv3 largely recover the same between-category structure in infant-view centroids but diverge more on THINGS centroids under the same category inventory.

\begin{table}[H]
  \centering
  \caption{Correlation between CLIP and DINOv3 cosine-distance RDMs (vectorized lower triangle, excluding the diagonal).}
  \label{tab:si_clip_dino_rdm}
  \begin{tabular}{llccc}
    \hline
    Category set & Dataset & Pearson $r$ & Spearman $\rho$ & Pairwise pairs \\
    \hline
    $N=129$ & BabyView & $0.907$ & $0.874$ & $8{,}256$ \\
    $N=129$ & THINGS     & $0.679$ & $0.487$ & $8{,}256$ \\
    $N=85$  & BabyView & $0.889$ & $0.839$ & $3{,}570$ \\
    $N=85$  & THINGS     & $0.702$ & $0.514$ & $3{,}570$ \\
    \hline
  \end{tabular}
\end{table}

\noindent All $p$-values were effectively zero at machine precision ($p<10^{-100}$ for the reported coefficients); we omit exact floating-point $p$ here.

\subsection*{SI 1.9. CLIP-threshold sensitivity for representational geometry and retained detections (N=129)}

To evaluate sensitivity to the CLIP image--text detection filter, we repeated the $N=129$ analysis over thresholds .26, .27, and .28 while keeping category scope and ordering fixed. For each threshold, BabyView category centroids were recomputed from retained exemplar crops, and RDM-based geometry metrics were recalculated against fixed THINGS centroids. We also tracked the total number of CLIP-filtered detections retained within the included category scope.

\begin{figure}[H]
  \centering
  \includegraphics[width=0.94\textwidth]{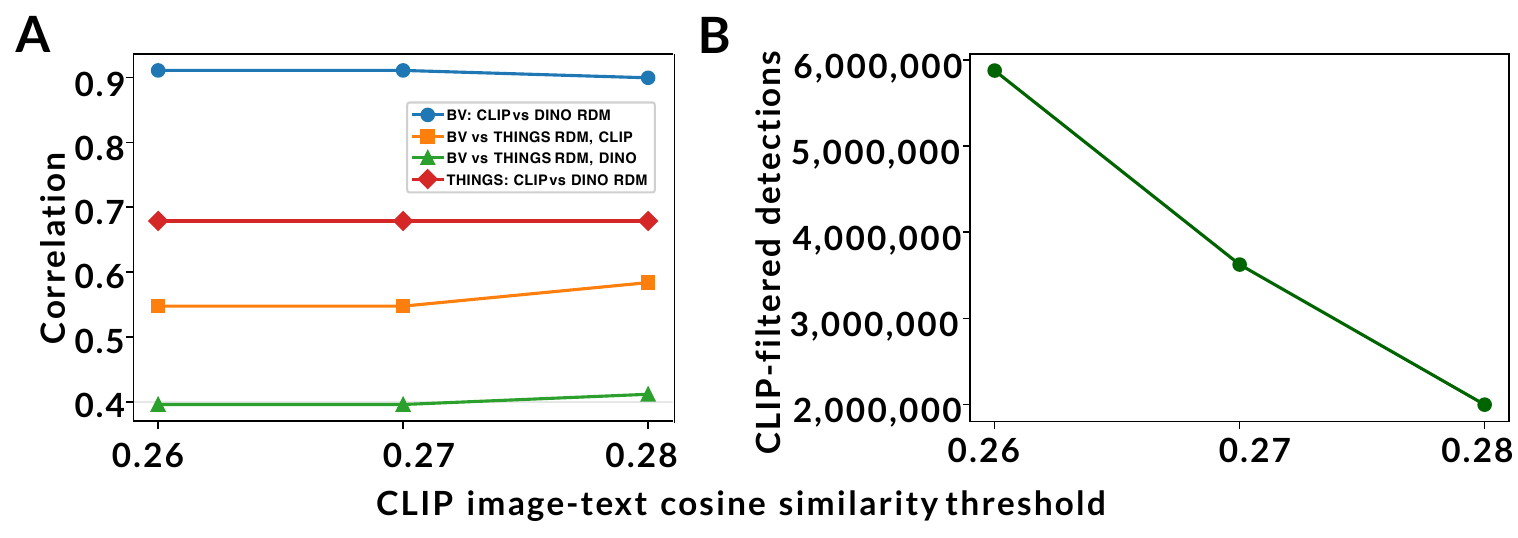}
  \caption{CLIP-threshold sensitivity analysis for the 129-category set. (A) Representational geometry correlations across thresholds, including BabyView--THINGS RDM correlations (CLIP and DINOv3) and within-dataset CLIP--DINO RDM agreement. (B) Total CLIP-filtered detections retained as threshold varies. The operating point used in main analyses ($t=0.27$) is marked for reference. Geometry metrics remain comparatively stable across .26--.28, while retained detections decrease monotonically as threshold increases.}
  \label{fig:si_clip_threshold_sensitivity_valid129}
\end{figure}

Figure~\ref{fig:si_clip_threshold_sensitivity_valid129} reports the two diagnostics used in the main robustness check: (A) representational geometry correlations across thresholds, and (B) total retained detections by category scope. Across this range, the geometry pattern is stable (small variation in RDM correlations) while retained-detection totals change monotonically with threshold, indicating that the main geometric conclusions are not driven by a single idiosyncratic cutoff at .27.

\end{document}